\documentclass{article}
\usepackage{arxiv}

\usepackage[utf8]{inputenc} 
\usepackage[T1]{fontenc}    
\usepackage{hyperref}       
\usepackage{url}            
\usepackage{booktabs}       
\usepackage{amsfonts}       
\usepackage{nicefrac}       
\usepackage{microtype}      
\usepackage{graphicx}
\usepackage{natbib}
\usepackage{doi}

\usepackage{longtable}
\usepackage{amsmath}
\usepackage{enumitem}

\title{A Scoping Review of Deep Learning for Urban Visual Pollution and Proposal of a Real-Time Monitoring Framework with a Visual Pollution Index}

\author{
\href{https://orcid.org/0000-0001-8091-511X}{
Mohammad Masudur Rahman},
{
Md. Rashedur Rahman},
{
Ashraful Islam},
{
Saadia B Alam},
{
M Ashraful Amin}
\\[2mm]
Center for Computational \& Data Sciences\\
Independent University, Bangladesh (IUB)\\
Dhaka, Bangladesh.\\
\texttt{\{masudurismsets, rashed, ashraful, saadiabinte, aminmdashraful\}@iub.ac.bd}
}

\date{}

\renewcommand{\shorttitle}{\textit{arXiv} Template}

\hypersetup{
pdftitle={A template for the arxiv style},
pdfsubject={q-bio.NC, q-bio.QM},
pdfauthor={David S.~Hippocampus, Elias D.~Striatum},
pdfkeywords={First keyword, Second keyword, More},
}

\begin{document}
\maketitle

\begin{abstract}
Urban Visual Pollution (UVP) has emerged as a critical concern, yet research on automatic detection and application remains fragmented. This scoping review maps the existing deep learning-based approaches for detecting, classifying, and designing a comprehensive application framework for visual pollution management. Following the PRISMA-ScR guidelines, seven academic databases (Scopus, Web of Science, IEEE Xplore, ACM DL, ScienceDirect, SpringerNatureLink, and Wiley) were systematically searched and reviewed, and 26 articles were found. Most research focuses on specific pollutant categories and employs variations of YOLO, Faster R-CNN, and EfficientDet architectures. Although several datasets exist, they are limited to specific areas and lack standardized taxonomies. Few studies integrate detection into real-time application systems, yet they tend to be geographically skewed. We proposed a framework for monitoring visual pollution that integrates a visual pollution index to assess the severity of visual pollution for a certain area. This review highlights the need for a unified UVP management system that incorporates pollutant taxonomy, a cross-city benchmark dataset, a generalized deep learning model, and an assessment index that supports sustainable urban aesthetics and enhances the well-being of urban dwellers.
\end{abstract}


\section{Introduction}
Humans generate approximately 2.2 billion tons of municipal solid waste annually \cite{un_data}. These pernicious activities are increasingly degrading the natural environment. Scientists named this subtle yet pervasive form of degradation as Urban Visual Pollution (UVP) \cite{nawaz2022visual}. It occurs when disorganized or intrusive elements misfit and disrupt the visual balance, diminishing the aesthetic quality of urban spaces. Examples include unusual billboards, excessive signage, misplaced waste dumps, overhead chaotic wiring, etc. \cite{portella2016visual}. Unlike other forms of pollution, visual pollution does not directly affect physical health, but it can have significant psychological impacts on the citizens. Many studies have shown that exposure to visually chaotic environments contributes to psychological fatigue, cognitive stress, and reduced sense of place attachment, ultimately degrading the quality of urban life and community well-being \cite{maclurcan1973}. However, visual pollution remains less explored in research. Traditional qualitative approaches rely on manual survey-based methods or subjective evaluations, which are time-consuming, error-prone, inconsistent, and not scalable or uniform \cite{nawaz2022visual}. As urban scenarios rapidly change with the construction of high-rise buildings, large billboards, and increasing demands for development infrastructure, automated monitoring for visual pollution is necessary. Without clear standards and reliable tools, it becomes challenging for planners to make data-driven decisions about maintaining visually balanced urban spaces, thereby ensuring better living conditions.

Visual pollution is closely linked to computer vision tasks, so using image-based technologies can help address it more effectively. Recent progress in computer vision and image analysis enables the automatic detection and classification of such problems. Modern deep learning techniques, such as Convolutional Neural Networks (CNN), can learn to identify visual anomalies in images and detect clutter, signage, or waste with significantly higher accuracy than manual methods \cite{silvy2024deep}. Earlier studies primarily employed statistical learning approaches, which often failed to capture the diverse range of visual pollution in cities, from small objects like litter to larger structures such as oversized billboards. Challenges exist in performing a comprehensive analysis to capture the complex forms of visual pollution \cite{szczepanska2019visual}. With recent advances in deep learning, many of these constraints can now be mitigated. Nevertheless, the deployment of deep learning techniques for visual pollution analysis is still nascent. Some studies often focus on specific pollution detection tasks, whereas there is a lack of a comprehensive approach for detecting and classifying a variety of pollutants. Some solutions are domain-specific, lacking a comprehensive understanding of how different model architectures perform under varying conditions. Furthermore, there is limited work on integrating these models into real-time, scalable frameworks that can support smart-city systems and public reporting. Hence, we intend to perform a comparative analysis of deep learning models to determine their suitability, limitations, and feasibility for deployment. This investigation, through rigorous comparative analysis, can help identify the best-performing architectures and a comprehensive dataset for the visual pollution solutions. Beyond model comparison, we believe it is equally important to develop a unified end-to-end framework that combines detection, contextual analysis, and real-time monitoring mitigation support. Following PRISMA-ScR guidance, this review aims to address the following research question:

\begin{enumerate}[label=RQ\arabic*.]
    \item What deep learning models and benchmark datasets have been used for visual pollution detection or classification?
    \item How can insights from existing studies support the development of a unified, deployable real-time framework?
    \item What are the challenges and prospects for implementing a real-time UVP monitoring framework?
\end{enumerate}

\subsection*{Organization of the paper}
We start the paper with an introduction and background analysis. This paper covers two aspects: a scoping review and a proposal for a real-time monitoring framework. The remainder of the paper is organized as follows: Section 2 explains the review method used in this study, employing the PRISMA-ScR technique. Section 3 discusses the results of the review. In Sections 4, 5, and 6, we explore the answers to the research questions. Section 4 examines state-of-the-art deep learning models and benchmark datasets, while Section 5 provides insights from existing studies. Section 6 addresses the challenges and prospects for implementing a UVP framework. Section 7 of the paper presents a proposal for a real-time mobile application framework to address visual pollution, concluding in Section 8.

\section{Review Method}
\label{sec:method}
This review followed the \textit{\emph{P}referred \emph{R}eporting \emph{I}tems for \emph{S}ystematic \emph{R}eviews and \emph{M}eta-\emph{A}nalyses \emph{E}xtension for \emph{S}coping \emph{R}eviews (PRISMA-ScR)} checklist \cite{tricco2018prisma}. A review protocol was developed before conducting the research. The protocol defined the research questions, eligibility criteria, search strategy, screening procedures, and data extraction plan.

\subsection{Eligibility Criteria}
\subsubsection{Inclusion criteria}
Articles were included in this review study if they met the following conditions:
\begin{itemize}
  \item Focused on urban visual pollution or visually disruptive urban elements.
  \item Used deep learning methods and provided empirical evaluation (accuracy, mAP etc.).
  \item Were peer-reviewed journal or conference publications.
  \item Was published in English.
\end{itemize}
\subsubsection{Exclusion criteria}
Articles were excluded if they:
\begin{itemize}
  \item Focused on non-visual forms of pollution (air, noise, marine waste).
  \item Used any other methods without deep learning.
  \item Provided surveys, citizen interviews, or conceptual discussions.
\end{itemize}

\subsection{Database sources}
Seven scholarly databases are searched: Scopus, Web of Science, ScienceDirect, SpringerNatureLink, IEEE Xplorer, ACM Digital Library, and Wiley Library. No filters were applied to the publication year.

\subsection{Search Strategy}
The search terms for each database were constructed using the following two core concept groups: 
\begin{enumerate}
\item \textit{Visual Pollution Terms:} "visual pollution," "urban visual disorder," "garbage pollution," "waste," "trash," etc. 
\item \textit{Deep Learning Terms:} "deep learning," "object detection," "classification," "index," "monitoring," etc.
\end{enumerate}

For each group, we connected the related terms with OR; the resulting concept blocks were then combined with AND to construct database-specific search strings. The generated phrase-level search strings, which included terms such as "visual pollution using deep learning," "visual pollution detection," "garbage classification," and "visual pollution index," were paired to create the search strings. The search strings are provided in Appendix \ref{append:search}.

\subsection{Selection and review process}
\begin{figure}[t]
  \centering
  {\includegraphics[width=0.8\textwidth]{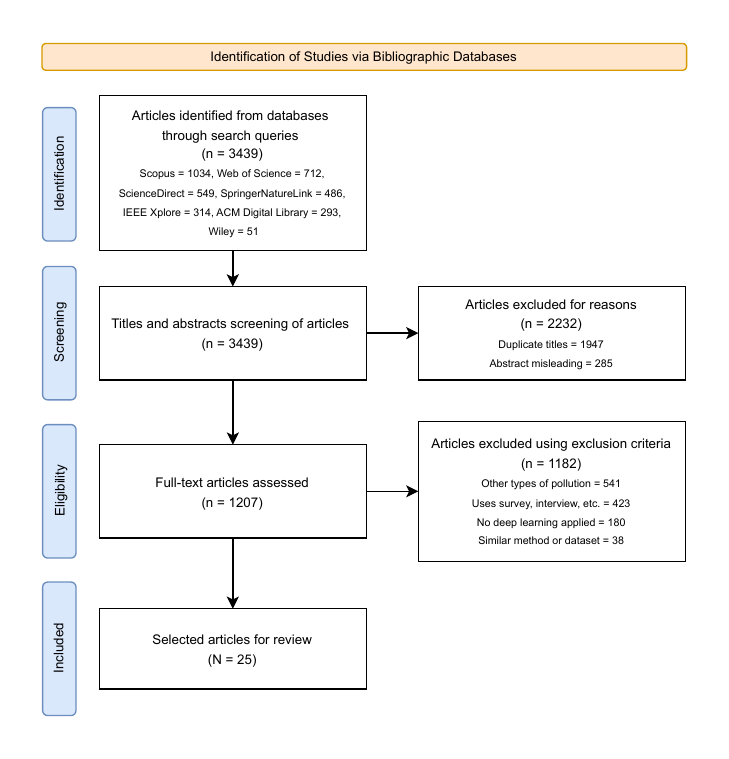}}
  \caption{PRISMA-ScR flow diagram for study selection process.}
  \label{fig:prisma-scr}
\end{figure}

The selection process for the study is shown in Fig. \ref{fig:prisma-scr}. Using the PRISMA-ScR flow diagram. Out of 3,439 articles found, 1,207 were retained after checking titles and abstracts, and 24 were finally nominated after reading the full texts. A commonly shared summary Excel sheet was prepared, listing the bibliographic information, study objective, types of visual pollution, and a summary of the deep learning model and evaluation metrics, with findings and limitations noted by the authors. At least two of the research team independently verified the authenticity of the article collection. Among the team, two researchers independently screened the titles and abstracts, followed by the full texts, using the established inclusion and exclusion criteria. Any differences in their decisions were discussed and resolved jointly, with assistance from a third reviewer when necessary.

\section{Results}
\subsection{Included Study characteristics}
We identified 26 publications through this review study, which address the detection, classification, and application of the visual pollution problem. The timeline reported in these publications spans from 2016 to 2025, indicating that recent developments in deep learning techniques have led to the effective solution of VP. To structure the synthesis of the literature, we organized the included studies into five themes that align with our research questions:
 
\begin{itemize}
  \item Detection- and classification-focused studies
  \item Application-level implementations
  \item Benchmark datasets
  \item Fragmented pollutant taxonomy
  \item Standardized visual pollution index
\end{itemize}

Among the 24 studies explored, 13 focus on detection and classification, while 14 involve the design and implementation of application frameworks, with most of them contributing to both categories. Detection, segmentation, and classification tasks are often difficult to distinguish in the context of pollutant identification, as they are closely coupled in computer vision pipelines and frequently co-occur within the same experimental setup. Some works consider visual pollution as a single-class object, whereas others view it as a multi-class or instance-level task. So, we keep them together while analyzing the studies. Appendix \ref{append:summary_class} provides a summary of detection and classification studies, while Appendix \ref{append:summary_app} represents application-level studies. These studies are conducted in various geographical locations, including developed countries such as the USA (n=2) and low-developed countries like Bangladesh (n=5), which indicates that VP is a global problem. Other countries included Poland (n = 1) and Ireland (n = 2), etc. The majority of the reviewed research (n = 23) employed different variations of YOLO (v3–v8), which is particularly effective at object detection. In addition, Faster R-CNN has been utilized in 11 studies, while EfficientDet was used in five, and transformer-based models, such as Swin Transformer, were implemented in another two studies. Only one study integrated explainable AI (XAI)~\cite{mazroa2024}. An overview of deep learning architectures is presented in Section \ref{models}. 

The categories of pollutants are diverse, with most studies focusing on billboards, roadside signs, and other similar sources of pollution. This implies the need to develop a coherent and universally accepted taxonomy of visual pollution types based on the scopes. Although several benchmark datasets are available for VP, they are limited in scope due to factors such as geo-location and pollution types. Most studies in this review relied on manually curated, task-specific datasets to train their models. A total of 18 studies developed customized datasets, while 14 utilized publicly available datasets (e.g., TACO, Place Pulse 2.0). A detailed analysis of the datasets is discussed in Section \ref{datasets}. In the review, we also intended to identify whether any Visual Pollution Index (VPI) formulations exist. We found that early works, particularly by Wakil et al. \cite{wakil2019assessment}, derive a VPI from paper-based audit scores, where different pollution categories (billboards, broken facades, cluttered signage, etc.) are rated and weighted using the Analytic Hierarchy Process. In this process, a group of 20 professionals enlist, classify, and rank the pollutants (visual pollution objects) with a rubric where they calculate a value between 0 and 100, called the visual pollution index (VPI) at any given location. The resulting scores are then mapped as choropleth or heat maps on web-based GIS dashboards to highlight hotspots and support intervention planning. Several other works later developed to quantify a score based on different criteria and proposed rankings or indices for the severity of visual pollution.
 
\subsection{State-of-the-Art DL Models and Benchmark Datasets}
Our first research question was \textit{RQ1: What are the recent advancements in detecting or classifying visual pollution using deep learning models and benchmark datasets?} To answer the question, this section analyzes the recent rise in deep learning methods for visual pollution, highlighting both advances in model architectures and the development of benchmark datasets.

\subsection{Deep learning enhances effectiveness}
\label{models}
Solving visual pollution using deep learning is rapidly advancing. This review focuses specifically on deep learning methods for the detection, classification, and application of visual pollution. Compared to the traditional system of manual assessment for visual pollution, deep learning techniques substantially enhance the automation, scalability, and decision-support capabilities of environmental monitoring systems. Deep learning techniques include convolutional neural networks (CNNs), recurrent neural networks (RNNs), and transformer-based models. From early developments of CNNs, these models have shown strong promise for classifying visual pollutants, as reflected in several studies, including \cite{ahmed2019}. The later development of RNN and its combination with CNN, such as Faster R-CNN for object detection and scene understanding in urban environments, is studied \cite{alzate2021}. Segmentation-based approaches such as Mask R-CNN have further driven the state of the art by enabling pixel-level delineation of pollutants in streetscapes \cite{majchrowska2022}. Deep learning–based object detection frameworks, such as YOLO and EfficientDet, have been widely adopted in visual pollution applications, with multiple YOLO variants being utilized in recent studies. These models balance between accuracy and inference speed, making them suitable for real-time deployment in urban monitoring systems. YOLO variations from v3 to v8 have been in several literature \cite{morera2020,tasnim2023,ptak2024mapping,alharbi2025}. Some recent studies have introduced explainability mechanisms for these deep learning models, aiming to reveal the formal reasoning behind the model decisions and to increase the interpretability of the solution \cite{mazroa2024}. DL-based techniques also have been integrated into application-level visual pollution solutions, including real-time monitoring, decision-support tools, and citizen-reporting platforms \cite{hu2023,campbell2019,mittal2016spotgarbage}. Studies like \cite{hossain2023end} demonstrate the use of mobile applications, while studies like \cite{titu2024,abozahhad2025SciRep} showcase Raspberry Pi-based applications. These developments demonstrate that deep learning integration can significantly enhance the detection and classification of visual pollutants, thereby expediting urban decision-making and contributing to the transformation of visual pollution from a manual, report-driven process into a real-time, immersive, and data-driven paradigm. Table \ref{tab:models} illustrates the overview of deep learning models used in detection, classification, segmentation, and application tasks.

\begin{table}[htbp]
\centering
\caption{Overview of deep learning models}
\label{tab:models}
\scriptsize
\begin{tabular}{p{0.12\linewidth} p{0.1\linewidth} p{0.22\linewidth} p{0.15\linewidth} p{0.28\linewidth}}
\toprule
\textbf{DL architectures} & \textbf{Task} & \textbf{Pipeline} & \textbf{Performance (best)} & \textbf{Limitations} \\
\midrule
VGG16/VGG19 &
Classification &
A total of 5 sequential convolution-max-pooling (ReLU) layers, 3 fully connected layers, a dropout rate of 50\%, and an L2 kernel regularizer are used &
85–90\% accuracy on custom datasets \cite{ahmed2019} &
Very high parameter count and FLOP that are not suitable for edge devices; high memory bandwidth; high energy consumption; latency too large on mobile (15–30 FPS) \\

ResNet50/ResNet101 &
Classification &
Residual learning with identity skip connections; bottleneck blocks in deeper variants &
$\sim$96\% accuracy on 30 classes \cite{jahan2025} &
Heavy model, may struggle with very small VP objects (distant billboards); accuracy drops after aggressive quantization/pruning \\

MobileNetV2/V3 &
Detection, Classification &
Depthwise separable conv, inverted residuals, linear bottlenecks; &
mAP 93\% \cite{alelaiwi2023vpp}; Detection acc. $\sim$95\%, classif. acc. $\sim$98\% \cite{campbell2019} &
Limited high-level context awareness: struggles with complex multi-class and dense scenes; aggressive down-scaling of input; careful tuning needed for small objects (potholes) \\

Faster R-CNN &
Detection &
Backbone using Region Proposal Network (RPN), then ROIAlign, then classification \& bounding box regressor head &
mAP=93\% \cite{alelaiwi2023vpp} &
Two-stage, high-latency architecture; even optimized GPU inference is expensive; complex to deploy on mobile; better as an offline system \\

Mask R-CNN &
Segmentation &
Faster R-CNN + parallel mask branch using deconvolutions &
mAP=97\% \cite{jahan2025} &
Heavy for edge-devices; mask head adds big memory \& time overhead; deployment to mobile only feasible via remote inference \\

EfficientDet (D0 to D7) &
Detection &
EfficientNet for detection tasks; weighted BiFPN; shared prediction head &
Acc: 97\% \cite{tasnim2023}, mAP @0.5: 0.86 \cite{rahmatulloh2025} &
Quite suitable for low-end phones; TF ecosystem heavier than lightweight YOLO implementations; export and optimization pipeline more complex but promising \\

YOLOv3 to YOLOv8 &
Detection &
Backbone architecture: YOLOv3=Darknet-53; v5=CSPDarknet; v6=EfficientRep; v7=E-ELAN; v8=Anchor-free decoupled head &
YOLOv7 and v8: mAP$\sim$0.96 in large datasets \cite{abozahhad2025SciRep, rahmatulloh2025,titu2024} &
Older variants are not much suitable but newer nano/small variants are a good fit; smallest custom MobileNet+SSD architecture; performance sensitive to training (augmentations, class imbalance handling); good fit for resource-constrained devices \\
\bottomrule
\end{tabular}
\end{table}

\subsection{Still scarcity of benchmark datasets}
\label{datasets}
Deep learning methods require sufficiently large and diverse datasets to perform well. The literature reveals a diverse set of datasets constructed for urban visual pollution (UVP) detection, classification, and segmentation. Many early datasets are relatively small, pollutant- and location-specific collections of waste materials or street elements captured for a narrow purpose. The TACO (Trash Annotations in Context) dataset has marked a pivotal improvement toward real, in-the-wild image collections with COCO-style annotations and instance segmentation masks, which significantly enhance object-level detection performance \cite{taco_data}. Numerous region-specific datasets have since emerged to capture the heterogeneity of urban environments. For example, Hossain et al. have compiled a dataset of Google Street View images for detecting waste, overhead wires, and posters in Dhaka, Bangladesh \cite{hossain2023end} while Ptak et al. focused on a dataset containing billboards and road signs in Poznan, Poland \cite{ptak2024mapping}. Datasets also vary by pollutant type like Tasnim et al. curated a dataset from textile industry environments \cite{tasnim2023}, Liu et al. concentrated on billboard advertisements \cite{liu2019}. Two studies considered a diverse set of pollutants: Jahan considered a dataset with 30 pollutant classes \cite{jahan2025}, whereas Rad et al. focused on 25 types \cite{rad2017}. Dataset sizes typically range from roughly 2,000 to 30,000 images, with most being manually curated and often lacking standardized validation. Although visual pollution is a global problem, many studies rely on location-specific datasets. For the Middle Eastern region, AlElaiwi et al. (2023) introduced the Urban Visual Pollution Real Images Benchmark, which covers eight pollutant categories, including graffiti, potholes, and overhead wires \cite{alelaiwi2023dataset1,alelaiwi2023dataset2}. Al Mazroa et al. (2025) expanded this effort with the Saudi Public Roads VP Dataset, comprising approximately 7,000 annotated images. Several datasets have been developed for Dhaka city, whereas countries such as China and Australia have a limited number of curated data collections. Other data collections, such as UAV Billboard Dataset \cite{ptak2024mapping} and the Illegal Advertising Dataset \cite{liu2020iad}, extend VP analysis to aerial and remote-sensing imagery. The Road Issue detection dataset combines several datasets into a single place and shows a considerable accuracy of 97\% using the YOLOv7 deep learning model \cite{disansa2025}. These datasets demonstrate the necessity for a multi-modal, context-rich, and geographically diverse dataset that enables real-time detection, segmentation, and visualization of visual pollutants. Most datasets remain class-imbalanced and inconsistently annotated, underscoring the need for standardized, globally representative benchmarks to fully realize the potential of deep learning–based visual pollution solutions. Table \ref{tab:datasets} gives us a glimpse of existing datasets.

\begin{table}[htbp]
\centering
\scriptsize
\caption{Overview of existing datasets}
\label{tab:datasets}
\begin{tabular}{p{0.17\linewidth} p{0.22\linewidth} p{0.12\linewidth} p{0.13\linewidth} p{0.22\linewidth}}
\toprule
\textbf{Datasets (year)} & \textbf{Domain / Classes} & \textbf{\# Images} & \textbf{Annotation type} & \textbf{Critiques} \\
\midrule
TACO (Trash Annotations in Context) (2020) \cite{proenca2020taco,taco_web} &
Litter/trash detection (urban, wild) &
$\sim$1,500 images, $\sim$4,784 annotations &
Instance segmentation masks (COCO-style) &
Most-used open trash dataset; limited to small objects; imbalanced classes, Small for in-the-wild deployment \\

Open Litter Map (2025) \cite{openlittermap_data} &
Crowdsourced geotagged litter / plastic pollution images from streets, parks and coasts &
8,200 photos and $\sim$28,000 labeled litter items &
Segmentation mask &
Regionally imbalanced and Europe-dominated; inconsistent annotations due to variable citizen inputs; lacks COCO-style bounding boxes or segmentation masks \\

Saudi Arabia Public Roads Visual Pollution Dataset (MOMRAH VP) (2023) \cite{alelaiwi2023dataset1,alelaiwi2023dataset2} &
Visual pollution on public roads (Saudi Arabia): Excavation barriers, potholes, dilapidated sidewalks &
34,460 RGB images &
Detection Labels using bounding boxes, classification label using captions (3 classes) &
Collection from real application, region-specific, relatively coarse class taxonomy \\

Urban Visual Pollution Dataset (UVPD) (2022) \cite{panigrahi2022} &
Outdoor scenes (graffiti, faded signage, potholes, garbage, construction road, broken signage, bad streetlight, bad billboard, sand on road, clutter sidewalk, unkept facade) &
9966 images, 19,950 labels &
Bounding box (11 type objects) &
Collected from moving vehicles, moderate size; class imbalance \\

pLitterStreet (2024) \cite{plitterstreet} &
Plastic litter in urban streets (from moving vehicles) &
$\sim$13,064 images ($\sim$78,333 instances) &
Bounding boxes &
Large, real-world scenes; limited to certain countries; domain narrow (plastic litter) \\

Dhaka City Visual Pollution Dataset (VP-Dhaka) (2023) \cite{hossain2023end} &
Billboards, Bricks, Construction Materials, Street Litter, Wires, and Towers &
1400 images (7 classes) &
Bounding boxes &
Combines urban and industrial pollution; limited daylight variation. \\

UAVBillboards Dataset (2024) \cite{ptak2024mapping} &
Drone  images of billboards &
1,361 images, 5,210 annotated objects, 3 classes &
Bounding boxes (YOLO format) &
Narrow domain (billboards only) \\

Illegal Advertising Dataset (IAD) (2020) \cite{liu2020iad} &
Unauthorized ad posters &
48,294 images, 13 classes &
Classification labels &
Some ambiguity in image classes \\

Garbage Classification Dataset (2024) \cite{mostafa2024garbage} &
Paper, cardboard, biological material, metal, plastic, green-glass, brown-glass, white-glass, clothes, shoes, batteries, and trash &
15,150 images, 12 classes &
Bounding box &
Household garbage \\

Billboard in Japanese Streetscapes () \cite{billboard_japanese} &
Billboard \& facade &
8,302 images &
Semantic segmentation masks &
Limited to Japanese context and billboard only \\

Place Pulse 2.0 (Extended) () \cite{placepulse} &
Urban physical disorder perception (litter, graffiti, decay) &
110,988 labeled street-view images &
Image-level + heatmap attention &
Large-scale; perceptual annotations; subjective ratings possible \\

Garbage Dataset () \cite{kunwar2024managing} &
Metal, glass, biological, paper, battery, trash, cardboard, shoes, clothes, plastic &
19,762 images &
Unannotated &
Clear and no-real background; indoor images\\

TrashNet () \cite{trashnet_data} &
Size classes: glass, paper, cardboard, plastic, metal, ttrash &
2527 images &
Unannotated &
In-the-wild images \\

TrashBox () \cite{venkat2022} &
Seven classes (glass, plastic, metal, e-waste, cardboard, paper, medical waste). &
17785 wastes &
Bounding box &
Includes medical and e-waste classes which are not included in any other existing dataset \\

Garbage In Images (GINI) Dataset (2016) \cite{mittal2016spotgarbage} &
Two classes: garbage and non-garbage &
2561 images &
Boundbox and one class label &
Bing searched images \\

UAVVaste () \cite{kraft_dataset} &
No class &
772 images, 3718 annotations &
Segmentation Mask &
Video dataset \\
\bottomrule
\end{tabular}
\end{table}

\subsection{Toward a Real-Time UVP Monitoring Framework}
This section explores answering the research question \textit{RQ2: How can comparative insights be synthesized into a unified framework that bridges research algorithms with deployable, real-world solutions?}

Urban Visual Pollution (UVP) research has made significant technical progress over the years, but this progress remains fragmented across isolated studies. Many works address one aspect of the larger problem, such as dataset creation, object detection or classification, segmentation, or the integration of a particular geolocation-dominated mobile application. However, no previous study provides an integrated pipeline that can function as a real-time working system. A unified framework is therefore necessary to consolidate these fragmented contributions into a coherent, scalable, and deployable solution. Such a system, therefore, requires some components:

\begin{enumerate}
  \item A clear problem definition and a taxonomy: Decide what counts as UVP in the framework: litter/waste, graffiti/defacement, signage and billboards/clutter, overhead utility cables, communication poles/towers, road and structural defects (potholes, broken sidewalks, facades), industrial/construction waste, etc. A level-wise hierarchical taxonomy is highly appreciated
  \item Annotation guidelines: COCO-style bounding boxes/masks can be implemented. The same label set and rules used across all cities/datasets are required for UVP as a global problem. Without this, models, datasets, and indexing scores won’t be comparable in the same context.
  \item Multi-source Data: A unified framework needs diverse, continuous visual input. The sources can include fixed CCTV, city cameras, UAV/drone surveys, citizen smartphone reporting, and more. Metadata is an important element in this process, including GPS coordinates, timestamp, camera ID, orientation, and weather/light conditions. So, a defined data collection pipeline is required.
  \item Deep learning model: It requires a stable, yet lightweight, deep learning model, as it will be implemented on edge or mobile devices. At the edge level, a small object detector like YOLOv8n is suitable for detecting pollution. At the server side, a high-performing segmentation model is required, for instance, semantic segmentation for area ratio and multi-label classification.
  \item A crucial gap across nearly all reviewed studies is the lack of a universal severity scoring system. Existing analyses mention area ratios or pollutant counts, but none provide a systematic, comparable metric. A Visual Pollution Index (VPI) that should standardize pollutant density, severity, recurrence frequency, and spatial sensitivity (hospitals, commercial zones). This index allows cities to prioritize cleaning, track longitudinal changes, and communicate conditions to the public, bridging the gap between algorithmic outputs and real-world decisions.
\end{enumerate}

The review studies show that many of the components needed for a complete visual-pollution monitoring framework already exist, but they are scattered across different problem settings. Some works focus on citizen- and mobile-based reporting. Mittal et al.~\cite{mittal2016spotgarbage} and Hossain et al.~\cite{hossain2023end} both develop mobile applications that enable users to capture images of garbage and local pollution, and send them along with location information. Abo-Zahhad et al.~\cite{abozahhad2025SciRep} and Rahmatulloh et al.~\cite{rahmatulloh2025} extend this idea to real-time edge devices, utilizing Raspberry Pi and similar hardware to continuously monitor garbage and various types of waste. Almalki and Algethami~\cite{almalki2025} add another layer by utilizing image captioning, enabling the system to not only detect visual pollution but also describe it in human-readable language, which is beneficial for automated reporting and integration with municipal systems. A group of studies works with street-view and geospatial imagery. Liu et al.~\cite{liu2019} detect illegal billboard advertising (FIBAD), while Campbell et al.~\cite{campbell2019} build an autonomous system to detect and classify traffic signs in Google Street View (GSV) images linked to GIS coordinates. Titu et al.~\cite{titu2024} combine Raspberry Pi–based robotic vision with GSV to identify streets and textile-related visual pollution. These works show how large-scale mapped imagery, when combined with geolocation, can create detailed spatial layers of visual pollutants over a city. Hu et al.~\cite{hu2023} propose UPDExplainer, a transformer-based framework that not only detects urban physical disorder but also explains which features contributed to the decision. Although it targets physical disorders more broadly than just visual pollution, the same idea—explaining why a scene is flagged—can be directly reused for VP. Almalki and Algethami proposed a captioning framework which contributes to interpretability by turning model outputs into natural language descriptions that planners and citizens can understand. When we put these strands together, a possible comprehensive framework for visual pollution becomes clearer \cite{almalki2025}. Mobile apps can serve as the crowd-sensing platform, edge devices provide continuous on-site monitoring, and GSV or GIS-based systems form a city-scale or area-based mapping system. However, these solutions are still largely isolated, as each targets a specific pollutant type, city, or hardware platform. Few works integrate a single loop that goes from detection $\rightarrow$ explanation $\rightarrow$ reporting $\rightarrow$ action $\rightarrow$ feedback. We can see the prospects of connecting these existing components into a unified pipeline, including crowd-sourced data collection (citizens, cameras, and GSV), deep learning–based detection and captioning, and direct interfaces to urban management systems for scheduling cleanup, enforcement, and long-term monitoring and management.

\subsection{Challenges and Prospects}
This section investigates the challenges and future prospects of using emerging technology to answer the question: \textit{RQ3: What are the challenges and prospects of implementing a comprehensive real-time framework for solving the visual pollution problem?}

\subsection{Challenges in Designing a Unified Framework for Visual Pollution}
Designing a unified framework for visual pollution is hard precisely because visual pollution itself is context-dependent. Here, we focus on the main challenges in designing a unified framework for addressing visual pollution.

Lack of a uniform taxonomy: Visual pollution is a growing concern, yet there is still no widely accepted taxonomy that clearly defines the conditions under which a visual disturbance should be classified as visual pollution. Such a taxonomy is based on geolocation, urban living standards, and the aesthetic and psychological perceptions of inhabitants. Urban planners identify billboards, illegal signage, and and chaotic facades as visual pollutants, whereas environmental scientists consider litter, waste dumps, and landfills as major pollutants, and others may identify road hazards, potholes, and guardrails as pollutants \cite{gitelman2019examination,vialytics2024dangersofpotholes}. Visual pollution is often a by-product of unplanned development that disrupts the aesthetic urban landscape~\cite{peyvandi2022location}. To reconcile these issues, a uniform taxonomy would provide the conceptual backbone for a unified framework, specifying which phenomena are in scope and how they are categorized. Although several works have been developed using different sets of pollutant types, they still lack a shared, coherent taxonomy. Table \ref{tab:taxonomy} represents a possible taxonomy of visual pollutants.

\begin{table}[t]
\centering
\scriptsize
\caption{List of visual pollutants (proposal for a uniform taxonomy)}
\label{tab:taxonomy}
\begin{tabular}{p{0.2\linewidth} p{0.4\linewidth} p{0.25\linewidth}}
\toprule
\textbf{Category} & \textbf{Visual pollutants} & \textbf{Referenced dataset} \\
\midrule
Structural and road defects &
Potholes, excavation barriers, broken or dilapidated sidewalks, road cracks, unkept facades, damaged walls, abandoned buildings, construction debris, sand on roads, high-rise buildings &
MOMRAH VP; UVPD; VP-Dhaka; RID; PP2 \\
Advertisement and signage &
Billboards, illegal advertisements, shop/store signboards, property signs, banners, posters, faded signage, traffic signs, public service boards, cluttered or overlapping signs &
UAVBillboards; IAD; UVPD; MOMRAH VP; BJS; VP-Dhaka; PP2 \\
Utility cables and wires &
Telephone and communication wires, overhead cables, dangling telecom wires, network towers, electric poles &
UVPD; MOMRAH VP; VP-Dhaka); PP2 \\
Litter and waste &
Street litter, garbage bags, trash bins overflow, plastic bottles, cans, paper, glass, metal, cardboard, organic waste, food waste, cigarette butts, electronic waste, scrap metal, rope, shoe, blister packs, batteries &
TACO; pLitterStreet; VP-Dhaka; GC Dataset; WC Dataset; Garbage Dataset; Trashnet; TrashBox; GINI; SmartTrashBin dataset; TIDY; WaDaBa; UAVVaste; LIV360SV Dataset; MOMRAH VP \\
Industrial waste &
Textile billboards and signage, dyeing factory waste, cloth dumps, dyeing materials, chemical residues &
VP-Dhaka; UVPD; MOMRAH VP \\
Graffiti and painted defacement &
Graffiti (tags, slogans, murals), vandalism on public surfaces, stickers on poles, painted-over signs &
UVPD; PP2; VP-Dhaka; MOMRAH VP; LIV360SV Dataset \\
\bottomrule
\end{tabular}
\end{table}

Developing benchmark dataset: Several visual pollution datasets have been proposed, but there is no widely accepted annotated benchmark that covers the full taxonomy of visual pollutants. Developing a reliable benchmark is therefore crucial for designing and evaluating a unified framework. Existing datasets typically compile images from Google Street View, crowdsourcing image collection, internet search, or collection from general-purpose image datasets, CCTV captures, drones images, satellite photos. Because urban visual pollution is a continuous and evolving problem, we argue for a data reinforcement strategy: newly collected images are periodically verified (e.g., by deep learning models and human reviewers) and then added to an open repository as additional training and evaluation data. Such a benchmark would reflect real-world conditions and support iterative improvement of the framework.

Data annotation problem: Data annotation is another crucial challenge, usually tackled via manual or semi-automated techniques. Common techniques of image annotation for visual pollution include: bounding box for pollutant detection, segmentation masks for spatial localization, and classification labels for categorizing pollution types. A better annotation can  affect deep learning model peroformance, hence, the reliability of any unified visual pollution framework.

Visual Pollution Indexing: Another open issue for the applicability of a unified framework is the absence of a standardized scale or index to quantify the severity of visual pollution. Other forms of pollution have well-established indicators, for example, the Air Quality Index (AQI) for ambient air pollution. A similar index is needed to characterize and compare the visual condition of different areas, track changes over time, and evaluate the impact of interventions. Nejadsadeghi et. al. showed the importance of having an index to asses littering activities by the citizen and educating citizens about their behaviors \cite{nejadsadeghi2024study}. Without such an index, it remains challenging to assess the severity of a given environmental impact from visual pollution. Wakil et. al. proposed a VPI manually auditing scores by professionals using pen and paper method \cite{wakil2019assessment}. Sowilska et al. introduced index of landscape disharmony (ILDH) that combines the aesthetic and ecological approach to landscape assessment \cite{sowilska2016}. Gupta et. al. uses another approach for indexing the severity of visual pollution \cite{gupta2024above}.

\subsection{Prospect of using LLM to solve visual pollution}
The recent rapid advancement and widespread adoption of Large Language Models (LLMs) have inspired us to think about the use of LLMs in solving visual pollution problems. Large Language Models (LLMs) and other AI technologies can be used for automated reporting and contextualization of pollution. Integrated with computer vision models (like YOLOx) which first detect visual pollutants (e.g., street litter, illegal billboards, dilapidated sidewalks, entangled wires) from images or video, LLM then uses the detection results to generate a detailed, contextual reports, which automates the whole process and reduces human incorporation. LLMs or Vision-Language Models (VLMs) can be utilized to analyze large datasets from various sources (sensors, satellite imagery, crowdsourced photos) to identify the phenomena of visual pollution and predict potential pollution hotspots to track environmental changes over time. It also can help the authorities and planners to prioritize actions, allocate resources efficiently, and make more informed, data-driven decisions for managing urban aesthetics and environment which may lead to developing regulatory strategies and policies for visual pollution.

\section{Framework design}
This section aims to provide a conceptual design of a comprehensive framework to address the visual pollution problem, assessing its feasibility by implementing or synchronizing current studies of methods based on their results. A layer-by-layer discussion is provided, focusing on the challenges of implementation.

We were motivated, for the first time, to design a holistic framework for solving urban visual pollution. 

\begin{figure}[t]
  \centering
  \caption{Proposed framework}
  \label{fig:proposed}
\end{figure}

Layer 1: Data Acquisition and Annotation
Data collection is a crucial part of the UVP framework. It should be a continuous process to collect data from various sources. Data sources may include heterogenous sources like, CCTV, UAV/drone images, video, street-view frames, smartphone captures of residents including metadata timestamp, device, and geolocation information. The annotation of data should is performed by expert systems. Use COCO-style schemas for annotation boxes/masks; bootstrap with a pre-trained auto-label, then curate with human verification. Some initiatives are promising to overcome this data collection phase as they crowdsource images of littering publicly using mobile apps \cite{openlittermap_data,taco_data}. However, we observe from the literature, many of available dataset are suffering from bounding box annotation and enriched classification labels. They provide only classification labels without bounding box and vice versa. For a robust design of a visual pollution framework, we require both of them for each image frames. Hence, a defined protocol can be utilized in this case. Majchrowska et al. proposed an useful protocol of two-stage system: litter localization and litter classification where the first stage uses object detector to search possible region of litter in an image, then the later stage determine the litter type \cite{majchrowska2022}. We utilize this method to concepualize the proposed framework in this study. To clarify, this stage does not overlap the actual deep learning model for detection task. 

Layer 2: Pollutant Detection Model
This layer trains a multi-class object detector to localize visual pollutants in an image. Annotated images from the data acquisition layer are utilized for training the model. Labels should follow a COCO-style schema and explicitly allow overlapping annotations for stacked/occluded items (n wall or poles, such as posters, crossing overhead wires); elongated objects are handled with small–object–oriented settings (extra low-stride heads). Selecting an appropriate object detection architecture is critical at this stage. As we aim to build a real-time, scalable system capable of detecting diverse forms of urban visual pollution, a one-stage model like YOLOv8 can provide the speed–accuracy balance necessary for deployment. YOLOv8 object detector offers several advantages as it does not require any predefined anchor boxes (preset shapes, sizes, and aspect ratios) to detect objects \cite{yaseen2024yolov8}. It has Cross-Stage Partial with Two-Way Fusion (C2f backbone) and a decoupled head that performs efficient feature extraction for small and elongated objects \cite{yaseen2024yolov8}. It also provides real-time inference that can be deployed on mobile devices and edge cameras. This training stage should uses spatiotemporal splits (different neighborhoods), mixed-precision and exponential moving average, with augmentations tuned for multi-scale, modest perspective, HSV jitter, motion blur/glare, and Copy-Paste to up-sample rare small objects, focal/quality focal losses etc. The model must evaluated using accuracy or class-wise precison-recall curves, mean avrage precison (mAP) and operational KPIs (latency, FPS, memory, params/FLOPs, energy per frame) across target device; and also robustness is sliced by lighting, weather, occlusion, scene type, and domain shift. Hence, we can build a scalable and sustained model for the framework.

Layer 3: Segmentation
Layer 3 aims to provide a server-side interpretation of city images by using segmentation to isolate visual pollutant regions in outdoor environments precisely. Unlike bounding-box detectors, the segmentation stage provides pixel-level delineation of litter, wires, broken infrastructure, and signage clutter, which is essential for fine-grained assessment. Segmentation utilized Intersection over Union (IoU), Dice coefficient, and boundary F-scores to ensure that edge precision and mask accuracy are quantified by evaluation, which is interpreted visually as binary masks or polygonal outlines, or pixel-count summaries, which are stored as structured spatial features. The resulting masks enable the quantification of the proportion of each image occupied by pollutants, allowing for area-based scoring rather than simple presence–absence detection. Several scaling factors, such as area percentage, regional intensity, and location distribution, can be calculated in the system that translates visual pollution into spatial indices. These indices serve as the foundation for severity indexing, hotspot identification, and threshold-based pollution assessment. Segmentation also supports spatial analytics such as locating pollutants in critical zones (e.g., roadsides, pedestrian pathways, building facades).

Layer 4: Pollutant Classification
After we identify potential visual pollutants in a scene, we still need to determine what they are and how significant they are. This stage gives meaning to raw detections by assigning human-understandable categories (e.g., illegal billboard, loose cable, litter) and turning them into interpretable signals for action (severity, urgency, and likely impact on residents). The purpose is threefold: (1) translate technical outputs into actionable recommendations; (2) distinguish look-alikes (e.g., permitted sign vs. illegal poster) so actions are fair and defensible; and (3) prioritize limited resources by estimating which cases most affect safety, aesthetics, or public sentiment. To achieve the goal, this stage requires a clear taxonomy that reflects local community standards. This layer can be performed in parallel with the previous segmentation layer, where the results can bring actionable decisions together, deciding which area and how much the pollutants are present, and what the types of pollutants are.

Layer 5: Reporting and Indexing
This stage converts model outputs into clear, place-based guidance for action. First, detection and segmentation, followed by classification, can be summarized by location (street segment, block, district) and time (daily/weekly trend), which produces a narrative of "what, where, and when" the visual pollution occurred. A novel Visual Pollution Index (VPI) is then computed to express overall severity on a scale of 0 to 100. 
The VPI typically blends several components: extent (counts/total area of pollutants), persistence (how often the issue recurs), exposure (traffic and footfall), proximity to sensitive locations (such as schools and clinics), and compliance status (e.g., illegal signage). The VPI can also provide a standardized metric to compare street- or city-wise severity of visual pollution and support longitudinal monitoring. We present a simple formula to compute VPI. Assume, D = Pollutant Density (Number of detected pollutants per unit; S = Severity/Area Ratio (Fraction of the frame occupied by pollutants (segmentation masks)); R = Recurrence Frequency (How often the exact location shows similar pollutants); and W = Spatial Context Weight (sensitive Area-based scaling, hospitals, Industrial regions, Residential areas). VPI is a weighted composite score based on these measures:

\begin{equation}
  \mathrm{VPI} = 100 \times (\alpha D + \beta S + \gamma R) \times W
\end{equation}

Where $\alpha$, $\beta$, $\gamma$ tunable weights (sum = 1). VPI Interpretation, like score (0–20), requires no intervention, (21–40) requires low routine maintenance, and eventually (81–100) indicates a critical situation of visual pollution. The VPI index  generates street-level and city-level heatmaps, severity ranking of neighborhoods, temporal trends (daily, weekly, seasonal) and predictive modeling of future pollution hotspots. For a real-time city monitoring system, the whole system should be light, fast, and consistent. The context of `Light' means it adds minimal delay on top of detection; `fast' means it keeps pace with live cameras, patrol routes, or drone flights; and `consistent' means it remains stable across neighborhoods, weather, and time of day. This consistency stems from two design choices, rather than technical schemes: a shared severity rubric that combines visibility, extent, and proximity to sensitive locations (such as schools and clinics); and a straightforward decision policy that links each label to an appropriate response (remove, repair, warn, fine, or monitor).

This framework has the potential to benefit the following concerns: 

Urban planning: We can achieve a holistic overview of a specific area for pollution by accumulating results or reports together. Identifying the areas with their levels of visual pollution can help urban planners to prioritize locations for interventions in order to improve the aesthetic quality of urban environments. Additionally, ranking of the pollutants can provide a more comprehensive understanding of the underlying causes and factors for pollution, thereby supporting more informed and effective urban planning decisions.

Pandemic control: Visual pollution is not only an aesthetic or psychological concern; it can also serve as a representative of environmental health risks relevant to pandemic prevention, such as dengue, in different countries. Many visual pollutants, such as unmanaged litter, illegal dumping, discarded containers, and clogged open drains, can create stagnant water, providing breeding habitats for Aedes mosquitoes and thereby increasing the likelihood of dengue transmission. Automated detection and spatial mapping of these elements using deep learning, hence a visual pollution framework can help to identify hotspots for source reduction, and environmental cleanup. In other pandemic contexts those that have external factors can be beneficial to this framework. These conditions can hinder compliance with hygiene, distancing, and vaccination campaigns. By localizing visual pollution, cities can prioritize where to deploy hygiene, and vaccination campaigns, redesign public spaces, and set up health related infrastructures in an integrated way.

\section{Conclusion}
This scoping review maps how deep learning is currently being applied to visual pollution with the long-term goal of enabling real-time monitoring and management towards a unified framework. Although several promising studies demonstrate the feasibility of detecting visual pollution, the literature remains fragmented, methodologically unsynchronized, and limited in scope. As urbanization instigates visual pollution growing at scale, a more unified design is needed: an open, multi-city framework with shared benchmarks, standardized taxonomies, interoperable datasets, and transparent reporting practices. Future work should embed human-in-the-loop mechanisms that involve city dwellers, AI practitioners, and policy makers for actionable enforcement and using spatiotemporal modeling to evaluate the impact of interventions over time. Deep learning for visual pollution is clearly promising, but to get the best of results, we need data, protocols, and deployment practices, and policy-aligned, human-centered systems that are designed to operate in real urban conditions.

\bibliographystyle{unsrtnat}
\bibliography{references}

\appendix
\section*{Appendices}
\section{Search Strategy}
\label{append:search}
\begin{table}[h]
\centering
\noindent\rule{\linewidth}{0.8pt}
\small
\textbf{Search Terms}\\
\noindent\rule{\linewidth}{0.5pt}
\text{T1} = \text{``visual pollution''}; \quad
\text{T2} = \text{``visual clutter''}; \\
\text{T3} = \text{``urban visual disorder''}; \quad
\text{T4} = \text{``waste''}; \\
\text{T5} = \text{``garbage''}; \quad
\text{T6} = \text{``litter''}; \\
\text{T7} = \text{``trash''}; \quad
\text{T8} = \text{``street pollution''}; \\
\text{T9} = \text{``detection''}; \quad
\text{T10} = \text{``classification''}; \\
\text{T11} = \text{``segmentation''}; \quad
\text{T12} = \text{``deep learning''}; \\
\text{T13} = \text{``computer vision''}; \quad
\text{T14} = \text{``management''}; \\
\text{T15} = \text{``index''}; \quad
\text{T16} = \text{``smart city''}; \quad
\text{T17} = \text{``monitoring''}. \\
\text{CONCEPT 1: T1 OR T2 OR T3 OR T4 OR T5 OR T6 OR T7 OR T8} \\ 
\text{CONCEPT 2: T9 OR T10 OR T11 OR T12 OR T13 OR T14 OR T15 OR T16 OR T17}\\
\noindent\rule{\linewidth}{0.5pt}
\textbf{Database Query Strings}\\
\vspace{5pt}
\begin{tabular}{p{0.28\linewidth} p{0.65\linewidth}}
\midrule
\textbf{Database} & \textbf{Query string} \\
\midrule
Scopus &
TITLE-ABS-KEY((CONCEPT 1) AND (CONCEPT 2)) \\
Web of Science &
TS = ((CONCEPT 1) AND (CONCEPT 2)) \\
ScienceDirect &
TITLE-ABSTR-KEY((CONCEPT 1) AND (CONCEPT 2)) \\
SpringerNatureLink &
(CONCEPT 1) AND (CONCEPT 2) \\
IEEE Xplore &
((CONCEPT 1) AND (CONCEPT 2)); Content Type = Journals/Conferences \\
ACM Digital Library &
((CONCEPT 1) AND (CONCEPT 2)) \\
Wiley Online Library &
(CONCEPT 1) \& (CONCEPT 2) \\
\bottomrule
\end{tabular}

\end{table}

\section{Appendix 2: Summary of the recent studies (Classification and detection task)}
\label{append:summary_class}

\scriptsize
\begin{longtable}{p{0.13\linewidth} p{0.12\linewidth} p{0.05\linewidth} p{0.15\linewidth} p{0.13\linewidth} p{0.13\linewidth} p{0.15\linewidth}}
\toprule
\textbf{Author (year)} & \textbf{Objectives} & \textbf{Country} & \textbf{Types of visual pollution} & \textbf{Dataset used} & \textbf{DL model used} & \textbf{Results} \\
\midrule
Ahmed et al. (2019) \cite{ahmed2019} &
Classification of visual pollutants &
Bangladesh &
(i) Billboards and signage, (ii) telephone and communication wires, (iii) network and communication towers, and (iv) street litter &
Custom dataset using Google Image Search (800 images) &
Basic 5-layer CNN (similar to VGG, as they didn’t mention any names) &
Training accuracy of 95\% and validation accuracy of 85\% showed the applicability of VP classification and overall environment management \\

Tasnim et al. (2023) \cite{tasnim2023} &
Automatic identification and classification of textile contaminants &
Bangladesh &
Three categories of textile visual pollutants: textile advertising billboards and signage, dyeing factory waste, and cloth dumps &
Google and Bing image search and manual capture of local textile images (1,709 search results, 350 local capture), after augmentation 800 images &
Regional proposal network (RPN) driven Faster R-CNN, YOLOv5, and EfficientDet (Object detection variation of EfficientNet) &
Training accuracy 97\% and validation accuracy 93\%. Performance descends EfficientDet $>$ YOLOv5 $>$ Faster R-CNN. First two models are considerable \\

Morera et al. (2020) \cite{morera2020} &
Comparative study of SSD and YOLO for detection task usder different variations &
Spain &
Outdoor advertising billboards &
Custom dataset of street ad panels from Internet and taking photos (5400-5295) &
SSD, YOLOv3 object detection &
Both models performed acceptably; YOLOv3 found better panel localization \\

Alzate et al. (2021) \cite{alzate2021} &
Identifying graffiti and their concentration zones using deep learning &
Colombia &
Graffiti (text-like tag, bock slogan/writing, mural) &
Google Street View images from Medellín City, STORM dataset (1022 images), Extended dataset (373) panoramas &
Adapted ASUM-DM with Faster R-CNN &
STORM AP 58.3\%, Extended dataset AP 69.14\%, zone concentration using heat map visualization \\

Jiguang Dai and Gu (2022) \cite{dai2022} &
Automatic extraction of store signboards from street view imagery under occlusion &
China &
Store signage (shop name boards on streets) &
Two self-labeled street-view signboard datasets: Chinese Text in the Wild (CTW) and Baidu, 3637 images &
OSO-YOLOv5 (modified YOLOv5 with enhanced C3, SPP modules) &
Achieved high accuracy signboard detection, boosting AP by 5–37.7\% \\

Lynch and Cuffe (2021) \cite{lynch2023} &
Automatic identification of signage on geotagged and timestamped streetscape images to flag illicit advertising &
Ireland &
Unlicensed ads/signs posted in public (on poles, walls) &
Manually captured 1,051 images of illegal advertising signage (Dublin) for fine-tuning &
SSD MobileNetV2 detection (with manual dataset) &
Property signs got 72.7\% precision, 87.5\% recall while Banners 66.7\% precision and 81.1\% recall \\

Wu et al. (2025) \cite{wu2025ioa} &
Present IOA-YOLO, a YOLO-based model enhanced for detecting illicit overhead cables &
China &
Unauthorized overhead cables (e.g., dangling telecom wires) &
City street images with annotated improper wiring &
IOA-YOLO (feature enhancement + dual perception), LTEM for object feature extraction, a GDPM to boost occlusion, and a HIDH for multi-scalability &
Real-time performance. 93.94\% precision and 88.17\% recall \\

Ptak and Kraft (2023) \cite{ptak2024mapping} &
Automatically detecting billboards using unmanned aerial vehicle (DJI Mini 2) images with GPS &
Poznan, Poland &
Billboards and road signs &
UAVBillboards dataset: First drone dataset for VP, 1361 images supplemented with spatial metadata, together with 5210 annotations &
YOLOv7, YOLOv8-L, and YOLOv8-X &
Mask mAP@0.5 0.472 to 0.927; Mask mAP 0.408 to 0.778, Web visualization \\

AlElaiwi et al. \cite{alelaiwi2023vpp} &
Used public road images to classify visual pollution &
Saudi Arabia &
Three categories: excavation barriers, potholes, and dilapidated sidewalks &
Saudi Arabia Public Roads Visual Pollution Dataset (34,460 RGB images) &
MobileNetV2, EfficientDet, Faster RCNN, Detectron2, and YOLOv5x &
Results: 89\% precision, 88\% recall, 89\% F1-score, and 93\% mAP \\

Alharbi et al. \cite{alharbi2025} &
Video classification using MoViNet to track public littering activities by vehicles and pedestrians + facial recog. and license plate detection &
Saudi Arabia &
Littering activities of a pedestrian or a vehicle &
Videos dataset: 276, Data augmentation &
Pedestrian face direction: OpenCV HaarCascade; Licence Plate: YOLOv8, SWAN system &
MoViNet and YOLOv8, achieved 99.5\% accuracy \\

Mazroa et al. (2024) \cite{mazroa2024} &
Computer Vision with Explainable AI for Visual Pollution Detection &
Saudi Arabia &
Graffiti, faded signage, potholes, garbage, construction road, broken signage, bad streetlight, bad billboard, sand on road, clutter sidewalk, unkept facade &
4400 images manually curated of 8 classes &
MCVXAI-VPD utilizes YOLOv5, combining CSP and SPP. Classification using BiLSTM; XAI model LIME &
MCVXAI-VPD achieves mAP 86.07, IoU 81.36, F1-score of 84.39 over YOLOv4, Faster-CNN, EfficientDet, SegFormer \\

Rad et al. \cite{rad2017} &
To quantifying glittering from street and sidewalks images &
Unknown &
25 types of waste &
Pretrained in ImageNet dataset and then captured 18,676 images to test &
OverFeat-GoogLeNet &
63.2\% of precision, 61.02\% of recall for the cigarette butts class. \\

Majchrowska et al. \cite{majchrowska2022,wimlds_trojmiasto} &
To detect and classify visual pollution in outdoor &
Global &
Bio waste, glass, metal and plastic, non-recyclable, paper, litter, and other &
Extended TACO, UAVVaste, TrashCan, TrashICRA, MJU-Waste, drinking-waste, and wade-ai &
Detection: EfficientDet, DETR, Mask R-CNN; Classification: EfficientNet-B2 &
Precision range for different datasets: 0.52 to 0.97, recall range: 0.51 to 0.97 F1-score range: 0.56 to 0.97 \\

Jahan et al. (2025) \cite{jahan2025} &
Real-time multi-class VP detection framework for urban waste &
Global &
Thirty distinct waste objects: battery, blister pack, bottle, glass, can, food waste, paper, plastic, rope \& strings, scrap metal, shoe, litter, cigarette, and so on &
TACO trash annotated dataset &
For detection: Mask R-CNN, SSD (Single Shot Multibox Detector) technology; For classification: ResNet-50, ResNet-18, MobileNetV3, and EfficientNetB5 &
Combinly built custom model: SSD MobileNet V2: mean Average Precision (mAP) of 97.15\% over 60 classes; SSD detects real-time objects with low-latency; ResNet-18 accuracy 96\% over 30 classes. Real-time lightweight adaptation proposed \\

Madhavi (2025) \cite{madhavi2025swinconvnext} &
A real-time garbage image classification &
Global &
12 classes: Paper, cardboard, biological material, metal, plastic, green-glass, brown-glass, white-glass, clothes, shoes, batteries, and trash &
Garbage Classification datasetl &
SwinConvNeXt: Swin Transformer with ConvNeXt, and a spatial attention mechanism &
Metric: 98.97\% accuracy, 98.42\% Precision, and 98.61\% Recall \\

\cite{dhang2024adsegnet} &
Identified the location of billboard in outdoor scenes &
Global &
Billboards, posters and screens having only advertisement placement &
ALOS dataset (Crowd-sourcing with Mapillary) 8065 images &
AdSegNet: fusion of VGG16 and SegNet &
Training accuracy: 98.58\%; Testing accuracy: 96.43\% \\
\bottomrule
\end{longtable}

\section{Summary of the recent studies (Application)}
\label{append:summary_app}

\scriptsize
\begin{longtable}{p{0.13\linewidth} p{0.12\linewidth} p{0.05\linewidth} p{0.15\linewidth} p{0.13\linewidth} p{0.13\linewidth} p{0.15\linewidth}}
\toprule
\textbf{Author (year)} & \textbf{Objectives} & \textbf{Country} & \textbf{Types of visual pollution} & \textbf{Dataset used} & \textbf{DL model used} & \textbf{Results} \\
\midrule
Hossain et al. (2023) \cite{hossain2023end} &
An end-to-end android application for pollution analysis and detection system &
Dhaka, Bangladesh &
Seven types of pollutants: billboards, street litters, constructions materials, bricks, wires, towers &
Google Street View 1400 images with GPS coordinates &
Faster R-CNN, YOLOv5, EfficientDet (Transfer learning) &
YOLOv5 with data augmentation with mAP 0.85, F1-score 0.84 \\

Liu et al. (2019) \cite{liu2019} &
Framework to detect illegal billboard advertising- FIBAD with data extraction &
China &
Billboard advertisements &
Custom dataset of billboard images in five domains (medical, food, real estate, public service etc.) with 2 classes (legal or illegal) &
Hybrid model with BERT for text classification, then image retrieval &
Extracted ad text and flagged legal or illegal billboards with an accuracy 87.5\% \\

Campbell et al. (2020) \cite{campbell2019} &
Build an autonomous system for detecting and classifying traffic signs on GSV images using GIS coordinates &
Australia &
Road traffic signs &
Google Street View images from 1191 GIS locations within 10m (2382 images) &
SSD MobileNet model &
Detection accuracy of 95.63\% and classification accuracy of 97.82\% \\

Hu et al. (2023) \cite{hu2023} &
Introduce UPDExplainer: a transformer framework to detect and explain urban physical disorder &
International (application case study of USA) &
Physical disorder (old/abandoned buildings, graffiti, litter, broken sidewalks) &
Re-annotated Place Pulse 2.0 dataset (street images with disorder labels), 110,988 street view images of 56 cities in 28 countries &
Swin Transformer (for UPD detection), explainable: factor identification and ranking using segmentation &
79.89\% detection accuracy for disorder scenes; mAP$\approx$75.5\%, highlighting causing disorder objects \\

Abo-Zahhad2025SciRep (2025) \cite{abozahhad2025SciRep} &
Real-time intelligent garbage monitoring and collection system using Raspberry PI &
Egypt &
Public trash bin, roadside trash, garbage bags: cardboard, glass, metal, paper, plastic, and trash &
TrashNet30 dataset (2527 images) and a custom dataset of 3000 images &
Two-stage cascaded classification YOLOv5 and YOLOv8 &
High precision (>90\%) in real-time litter detection. Accuracy from 92\% to 96\% \\

Almalki \& Algethami (2025) \cite{almalki2025} &
A framework for Image captioning to detect, classify, and describe visual pollution and reporting &
Saudi Arabia &
Potholes, excavation barriers, and dilapidated sidewalks, broken/faded signs, roadside trash, etc. &
Saudi Arabia Public Roads Visual Pollution Dataset (31.795 images with 41,804 labels) and Urban Visual Pollution Dataset (9966 images with 19,950 labels) &
YOLOv5 and EfficientDet ensemble for VP detection, BLIP-2 for image captioning based on the content of images &
Ensemble achieved mAP 0.95, precision 0.91, recall 0.95 (F1$\approx$0.93), outperforming individual models. Auto-generated captions ~80\% accurate \\

Rahmatulloh et al. (2025) \cite{rahmatulloh2025} &
Real-time identification of various types of waste (WasteInNet) &
Indonesia &
Urban and rural outdoor waste: Electronic, glass, metal, organic, paper, plastic &
Combined multiple open datasets (e.g. Waste cls., garbage cls, Cables-hvidx, Amicable, Batteries dataset), 1210 images &
EfficientDet, YOLOv7 and YOLOv7-tiny (with data augmentation and optimization) &
Precision ~0.801, mAP@0.5 = 0.868 on test set (7 litter categories) \\

Titu et al. \cite{titu2024} &
A Raspberry Pi robotic vision system Google Street View to identify streets and textile-based VP &
Dhaka, Bangladesh &
communication towers, electric cables, construction materials, street litter, billboards, cloth dumps, dyeing materials, bricks &
Google and Bing search: 1940 photos; Textile: 598 photos; Augmented: 3018; Total: 5653 &
Faster SegFormer, YOLOv5, YOLOv7, and EfficientDet &
YOLOv5 and YOLOv7 models achieved 98\% and 92\% detection accuracy. YOLOv5 implemented in Raspberry Pi \\

Mittal et al. \cite{mittal2016spotgarbage} &
A mobile app: SpotGarbage to track and report garbage locations using geo-tagged image from citizens and their neighborhood &
Random &
garbage and non-garbage &
Bing Image Search API, 2561 images (GINI dataset) &
GrabNet, pre-trained AlexNet &
Accuracy 87.69, sensitivity 83.96, specificity 90.06 \\

\cite{kako2024quantification} &
App for visualizing the location and abundance of street litter (the number of litter pieces) &
Japan  &
Nine classes: cans, plastic bags, plastic bottles, cigarette butts, cigarette boxes, and sanitary masks, food packings, others &
Crowd-scouring image collection, 637+756 images &
YOLOv5, COCO annotation &
precision and recall > 75\% \\
\bottomrule
\end{longtable}

\end{document}